%% file: root.tex
\documentclass{bmvc2k}

\usepackage{url}
\usepackage{amsmath,amsfonts,amssymb,mathrsfs}
\usepackage{verbatim}
\usepackage{mathrsfs}
\let\chapter\section
\usepackage[ruled,linesnumbered, noend]{algorithm2e}
\usepackage{wrapfig}
\usepackage{graphicx} 
\usepackage{xspace}
\usepackage{subfigure}

\usepackage{times}
\usepackage{multirow,multicol}
\usepackage{color}

\title{Robust  6D Object Pose Estimation\\ with Stochastic
  Congruent Sets}

\addauthor{Chaitanya Mitash}{cm1074@rutgers.edu}{1}
\addauthor{Abdeslam Boularias}{ab1544@rutgers.edu}{1}
\addauthor{Kostas Bekris}{kb572@rutgers.edu}{1}

\addinstitution{
 Department of Computer Science\\
 Rutgers University\\
 New Jersey, USA
}

\runninghead{Mitash, Boularias, Bekris}{Robust Pose Estimation}

\input{notation}

\begin{document}

\maketitle

\begin{abstract}
\input{00_abstract}
\end{abstract}

\section{Introduction}
\label{sec:intro}
\input{01_introduction}

\section{Related Work}
\label{sec:related}
\input{02_relatedwork}

\vspace{-0.3cm}
\section{Approach}
\label{sec:approach}
\input{04_approach}

\section{Evaluation}
\label{sec:evaluation}
\input{05_evaluation}

\section{Discussion}
\label{sec:conclusion}
\input{06_conclusion}

\bibliography{physics_perception}
\end{document}

%% file: notation.tex
\newcommand{\procs}{{\tt StoCS}}
\newcommand{\stocs}{{\tt StoCS}}
\newcommand{\super}{{\tt Super4PCS}}
\newcommand{\vpcs}{{\tt V4PCS}}
\newcommand{\posecnn}{{\tt PoseCNN}}
\newcommand{\ycbvideo}{{\tt YCB-Video}}
\newcommand{\ycb}{{\tt YCB}}
\newcommand{\apc}{{\tt APC}}
\newcommand{\icp}{{\tt ICP}}
\newcommand{\fcn}{{\tt FCN}}
\newcommand{\cnn}{{\tt CNN}}
\newcommand{\rgbd}{{\tt RGB-D}}
\newcommand{\rgb}{{\tt RGB}}

\newcommand{\object}{O}
\newcommand{\model}{M}
\newcommand{\trans}{T}
\newcommand{\image}{I}
\newcommand{\segment}{S}
\newcommand{\base}{B}
\newcommand{\congsuperset}{\mathcal{U}}
\newcommand{\congset}{U}

\newcommand{\fweight}{w_{k}}
\newcommand{\pixel}{p}
\newcommand{\mpoint}{m}
\newcommand{\closest}{s^*}
\newcommand{\pixelprob}{\pi(\pixel_i \rightarrow \object_k)}
\newcommand{\potential}{\phi}

\DeclareMathOperator*{\argmax}{arg\,max}

{\end{list}}

%% file: 00_abstract.tex
\noindent Object pose estimation is frequently achieved by first
segmenting an {\tt RGB} image and then, given depth data, registering
the corresponding point cloud segment against the object's 3D
model. Despite the progress due to \cnn s, semantic segmentation
output can be noisy, especially when the \cnn\ is only trained on
synthetic data. This causes registration methods to fail in estimating
a good object pose. This work proposes a novel stochastic optimization
process that treats the segmentation output of \cnn s as a confidence
probability. The algorithm, called Stochastic Congruent Sets (\stocs),
samples pointsets on the point cloud according to the soft
segmentation distribution and so as to agree with the object's known
geometry. The pointsets are then matched to congruent sets on the 3D
object model to generate pose estimates. \stocs\ is shown to be robust
on an \apc\ dataset, despite the fact the \cnn \ is trained only on
synthetic data. In the \ycb\ dataset, \stocs\ outperforms a recent
network for 6D pose estimation and alternative pointset matching
techniques.


%% file: 01_introduction.tex
Accurate object pose estimation is critical in the context of many
tasks, such as augmented reality or robotic manipulation. As
demonstrated during the the Amazon Picking Challenge ({\tt
APC}) \cite{Correll:2016aa}, current solutions to 6D pose estimation
face issues when exposed to a clutter of similar-looking objects in
complex arrangements within tight spaces.

\begin{figure*}
\begin{center}
\includegraphics[width=\textwidth]{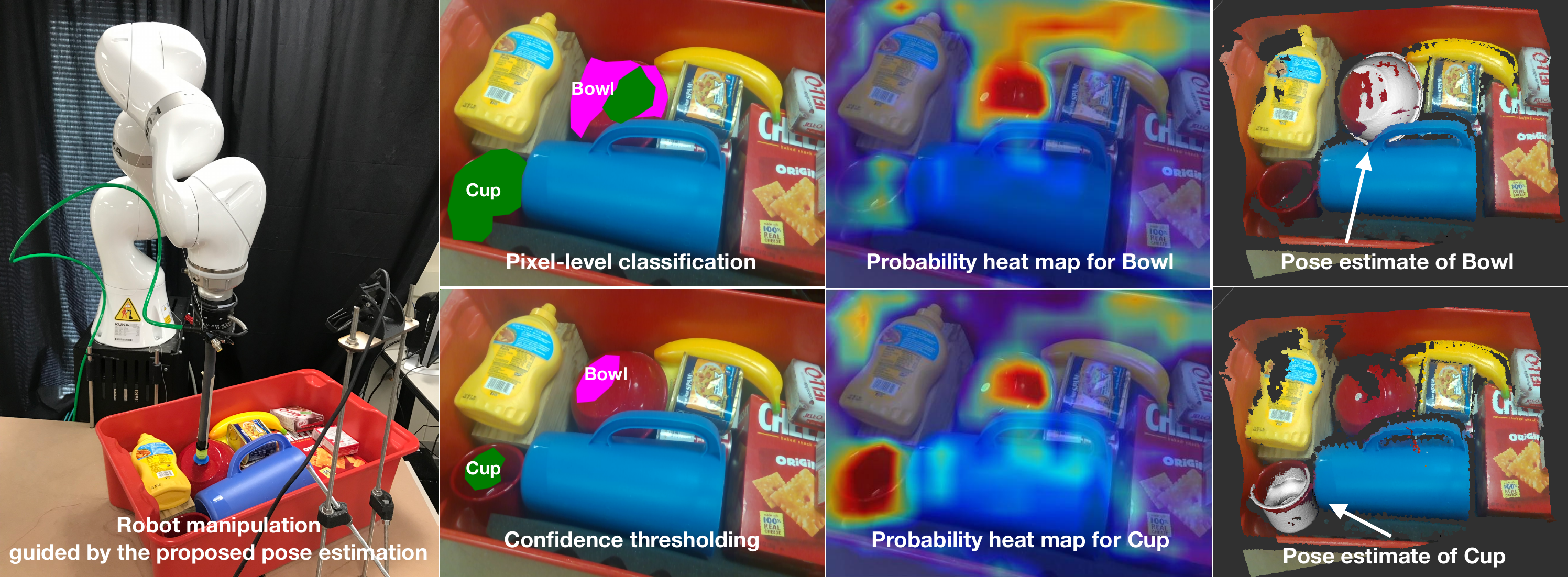}
\end{center}
\vspace{-.2in}
   \caption{ (a) A robotic arm using pose estimates from \stocs\ to
   perform manipulation. (b) Hard segmentation errors adversely affect
   model registration. (c) Heatmaps showing the continuous probability
   distribution for an object. (d) Pose estimates obtained by \stocs.}
\vspace{-.2in}
\label{fig:intro_image}
\end{figure*}

Solving such problems frequently involves two sub-components,
image-based object recognition and searching in SE(3) to estimate a
unique pose for the target object. Many recent
approaches \cite{Princeton, hernandez2016team, shelhamer2016fully,
ren2015faster} treat object segmentation by using a Convolutional
Neural Network (\cnn), which provides a per-pixel classification. Such
a hard segmentation approach can lead to under-segmentation or
over-segmentation, as shown in Fig. \ref{fig:intro_image}.

Segmentation is followed by a {\tt 3D} model alignment using point
cloud registration, such as \icp\ \cite{icp}, or global search
alternatives, such as 4-points congruent sets
(4-PCS) \cite{aiger20084, mellado2014super}. These methods operate
over two deterministic point sets $\segment$ and $\model$. They sample
iteratively, a base $\base$ of 4 coplanar points on $\segment$ and try
to find a set of 4 congruent points on $\model$, given geometric
constraints, so as to identify a relative transform between $\segment$
and $\model$ that gives the best alignment score. The pose estimate
from such a process is incorrect when the segment is noisy or if it
does not contain enough points from the object.

The key observation of this work is that \cnn\ output can be seen
as a probability for an object to be visible at each pixel. These
segmentation probabilities can then be used during the registration
process to achieve robust and fast pose estimation. This requires
sampling a base $\base$ on a segment, such that all points on the base
belong to the target object with high probability. The resulting
approach, denoted as ``Stochastic Congruent Sets'' (\stocs), achieves
this by building a probabilistic graphical model given the obtained
soft segmentation and information from the pre-processed geometric
object models. The pre-processing corresponds to building a global
model descriptor that expresses oriented point pair
features~\cite{drost2010model}. This geometric modeling, not only
biases the base samples to lie within the object bound, but is also
used to constrain the search for finding the congruent sets, which
provides a substantial computational benefit.

Thus, this work presents two key insights: 1) it is not necessary to
make hard segmentation decisions prior to registration, instead the
pose estimation can operate over the continuous segmentation
confidence output of \cnn s. 2) Combining a global geometric
descriptor with the soft segmentation output of \cnn s intrinsically
improves object segmentation during registration without a
computational overhead.

\stocs\ is first tested on a dataset of cluttered real-world scenes by 
using the output of an \fcn\ that was trained solely on a synthetic 
dataset. In such cases, the resulting segmentation is quite
noisy. Nevertheless, experiments show that high accuracy in pose
estimation can be achieved with \stocs. The method has also been
evaluated on the \ycb\ object dataset~\cite{xiang2017posecnn}, a
benchmark for robotic manipulation, where it outperforms modern
pointset registration and pose estimation techniques in accuracy. It is
much faster than competing registration processes, and only slightly
slower than end-to-end learning.

%% file: 02_relatedwork.tex
A pose estimation approach is to match feature points between textured
3D models and images \cite{Lowe:1999aa, Rothganger:2006aa,
Collet:2011aa}. This requires textured objects and good lighting,
which motivates the use of range data. Some range-based techniques
compute correspondences between local point descriptors on the scene
and the object model. Given correspondences, robust
detectors \cite{ballard1981generalizing, fischler1981random} are used
to compute the rigid transform consistent with the most
correspondences. Local descriptors \cite{tombari2010unique,
rusu2009fast, johnson1999using} can be used but they depend on local
surface information, which is heavily influenced by resolution and quality of
sensor and model data \cite{aldoma2012tutorial}. The features are
often parametrized by the area of influence, which is not trivial to
decide.


A way to counter these limitations is to use \emph{oriented point-pair
features} \cite{drost2010model} to create a map that stores the model
points that exhibit each feature. This map can be used to match the
scene features and uses a fast voting scheme to get the object pose.
This idea was extended to incorporate color \cite{choi20123d},
geometric edge information \cite{drost20123d} and visibility
context \cite{birdal2015point, kim20113d}. Recent work \cite{hinterstoisser2016going} 
samples scene points by reasoning
about the model size. Point-pair features have been criticized for
performance loss in the presence of clutter, sensor noise and due to
their quadratic complexity.

\emph{Template matching}, such as {\tt
LINEMOD} \cite{hinterstoisser2012model, Hinterstoisser:2012aa},
samples viewpoints around a {\tt 3D CAD} model and builds templates
for each viewpoint based on color gradient and surface normals. These
are later matched to compute object pose. This approach tends not to
be robust to occlusions and change in lighting.

There are also end-to-end pose estimation pipelines \cite{xiang2017posecnn, kehl2017ssd} 
and some approaches based on learning for predicting 3D object 
coordinates in the local model frame \cite{brachmann2014learning,
Tejani:2014aa, krull2015learning}. A recent
variant \cite{michel2017global} performs geometric validation on these
predictions by solving a conditional random field. Training for such
tasks requires labeling of 6D object poses in captured images, which
are representative of the real-world clutter. Such datasets are
difficult to acquire and involve a large amount of manual effort.
There are efforts in integrating deep learning with global search for
the discovery of poses of multiple objects \cite{Narayanan:2016aa} but
they tend to be time consuming and only deal with 3D poses.

Many recent pose estimation techniques \cite{Princeton,
hernandez2016team, mitash2017improving} integrate \cnn s for
segmentation with pointset registration such
as Iterative Closest Points (\icp) \cite{icp} and its
variants \cite{Rusinkiewicz:2001aa, Mitra:2004aa, Segal:2009aa,
Bouazix:2013aa, Srivatsan:2017aa}, which typically require a good
initialization. Otherwise, registration requires finding the best
aligning rigid transform over the {\tt 6-DOF} space of all possible
transforms, which are uniquely determined by 3 pairs of
(non-degenerate) corresponding points. A popular strategy is to invoke
{\tt RANSAC} to find aligning triplets of point
pairs \cite{Irani:1996aa} but suffers from a frequently observable
worst case $O(n^3)$ complexity in the number $n$ of data samples,
which has motivated many extensions \cite{Gelfand:2005aa,
Cheng:2013aa}.

The {\tt 4PCS} algorithm \cite{aiger20084} achieved $O(n^2)$
output-sensitive complexity using 4 congruent points basis instead of
3. This method was extended to \super\ \cite{mellado2014super}, which
achieves $O(n)$ output-sensitive complexity. Congruency is defined as
the invariance of the ratios of the line segments resulting from the
intersections of the edges connecting the 4 points. There are 2
critical limitations: (a) The only way to ensure the base contains
points from the object is by repeating the complete registration
process with several initial hypotheses; (b) The number of congruent
4-points in the model can be very large for certain bases and object
geometries, which increases computation time.

The current work fuses the idea of global geometric modeling of
objects along with a sampling-based registration technique to build a
robust pose estimator. This fusion can still enjoy the success of deep
learning but also remain immune to its limitations.

%% file: 04_approach.tex
Consider the problem of estimating the 6D poses of $N$ known objects
$\{ \object_1, \ldots, \object_N \}$, captured by an {\tt RGB-D}
camera in an image $\image$, given their 3D models
$\{ \model_1, \ldots, \model_N \}$. The estimated poses are returned
as a set of rigid-body transformations
$\{ \trans_1, \ldots, \trans_N \}$, where each $T_i = (t_i, R_i)$
captures the translation $t_i \in R^3$ and rotation $R_i \in SO(3)$ of
object model $\model_i$ in the camera's reference frame. Each model is
represented as a set of 3D surface points sampled from the object's
CAD model by using Poisson-disc sampling.


\subsection{Defining the Segmentation-based Prior}
\vspace{-.1in}

The proposed approach uses as prior the output of pixel-wise
classification. For this purpose, a fully-convolutional neural
network \cite{shelhamer2016fully} is trained for semantic segmentation
using {\tt RGB} images annotated with ground-truth object classes. The
learned weights of the final layer of the network $\fweight$ are used
to compute $\pi( \pixel_i \rightarrow \object_k )$, i.e., the
probability pixel $\pixel_i$ corresponds to object class
$\object_k$. In particular, this probability is defined as the ratio
of the weight $\fweight[p_i]$ over the sum of weights for the same
class over all pixels $\pixel$ in the image $\image$:

\vspace{-.25in}
\begin{eqnarray}
\pixelprob =  \frac{\fweight[\pixel_i]}{\sum_{\pixel \in \image}\fweight[\pixel]}.
\label{eq:soft_seg}
\end{eqnarray}
\vspace{-.2in}

These pixel probabilities are used to construct a point cloud segment $\segment_k$ for each object $\object_k$ by liberally accepting pixels
in the image that have a probability greater than a positive threshold $\epsilon$ and projecting them to the {\tt 3D} frame of the camera. The segment $\segment_{k}$ is accompanied by a probability distribution $\pi_{k}$ for all the points $p \in \segment_k$, which is defined as follows:

\vspace{-.25in}
\begin{eqnarray}
\segment_{k} \gets \{\pixel_i \mid \pixel_i \in I \wedge \pixelprob > \epsilon\}.\\
\pi_k(p) = \frac{\pixelprob}{\sum_{\forall q \in \segment_k} \pi(q \rightarrow O_k)}.
\label{eq:soft_seg2}
\end{eqnarray}
\vspace{-.2in}

Theoretically, $\epsilon$ can be set to 0, thus considering the entire
image. In practice, $\epsilon$ is set to a small value to avoid areas
that have minimal probability of belonging to the object.

\vspace{-.1in}
\subsection{Congruent Set Approach for Computing the Best Transform}
\vspace{-.1in}

The objective reduces to finding the rigid transformation that optimally aligns the model $\model_k$ given the point cloud segment
$\segment_{k}$ and the accompanying probability distribution $\pi_{k}$. To account for the noise in the extracted segment and the unknown overlap between the two pointsets, the {\it alignment objective} $\trans_{opt}$ is defined as the matching between the observed segment $\segment_{k}$ and the transformed model, weighted by the probabilities of the pixels. In particular:

\vspace{-.25in}
\begin{eqnarray*}
\trans_{opt} = \argmax_{\trans}\sum_{\mpoint_i \in \model_k}f( \mpoint_i, \trans,
\segment_k), \textrm{ where }\\
\ \ \ f(\mpoint_i, \trans, \segment_k ) =
\begin{cases}
\pi_k(\closest), \text{if} \mid \trans(\mpoint_i) - \closest \mid < \delta_s \wedge T(N(\mpoint_i)) \cdot N(\closest) > \delta_n \cr 
0, \text{otherwise.}
\end{cases}
\end{eqnarray*}
\vspace{-.2in}

\noindent where $\closest$ is the closest point on segment $\segment_k$
to model point $m_i$ after $m_i$ is transformed by $\trans$; N(.) is
the surface normal at that point; $\delta_s$ is the acceptable
distance threshold and $\delta_n$ is the surface normal alignment
threshold. Algorithm~\ref{alg:alg1} explains how to find
$\trans_{opt}$.

\input{compute_best_transform}

The proposed method follows the principles of randomized alignment
techniques and at each iteration samples a base $\base$, which is a
small set of points on the segment $\segment_k$. The sampling process
also takes into account the probability distribution $\pi_k$ as well
as geometric information regarding the model $\model_k$. To define a
unique rigid transform $\trans_i$, the cardinality of the base should
be at least three. Nevertheless, inspired by similar
methods \cite{aiger20084, mellado2014super}, the accompanying
implementation samples four points to define a base $\base$ for
increased robustness. The following section details the base selection
process.

For the sampled base $\base$, a set $\congsuperset$ of all similar or
congruent 4-point sets is computed on the model point set $\model_k$,
i.e., $\congsuperset$ is a set of tuples with 4 elements. For each of
the 4-point sets $\congset_j \in \congsuperset$ the method computes a
rigid transformation $\trans$, for which the optimization cost is
evaluated, and keeps track of the optimum transformation
$\trans_{opt}$. In the general case, the stopping criterion is a large
number of iterations, which are required to ensure a minimum success
probability with randomly sampled bases. In practice, however, the
approach stops after a maximum predefined runtime is reached.

\begin{figure*}[t]
\begin{center}
\fbox{\includegraphics[width=\textwidth]{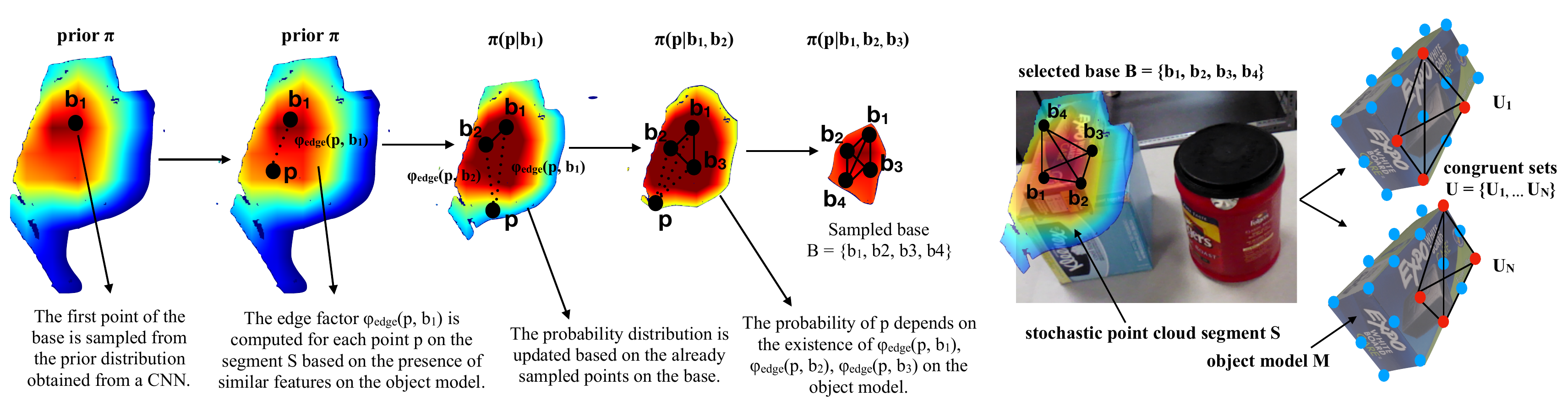}}
\end{center}
\vspace{-.2in}
\caption{A description of the stochastic optimization process for
extracting the base $\base = \{ b_1, b_2, b_3, b_4\}$ so that it is
distributed according to the stochastic segmentation and in accordance
with the object's known geometry. The base is matched against
candidate sets $\congset = \{ \congset_1$, \ldots, $\congset_N\}$ of 4
congruent points each from the object model $\model$.}
\vspace{-.2in}
\label{fig:base_sampling}
\end{figure*}

\vspace{-.1in}
\subsection{Stochastic Optimization for Selecting the Base}
\vspace{-.1in}

The process for selecting the base is given in Alg.~\ref{alg:alg2} and
highlighted in Fig. \ref{fig:base_sampling}. As only a limited number
of bases can be evaluated in a given time frame, it is critical to
ensure that all base points belong to the object in consideration with
high probability.  Using the Hammersley-Clifford factorization, the
joint probability of points in $\base = \{b_{1}, b_{2}, b_{3},
b_{4} \mid b_{1:4} \in S_k\}$ belonging to $\object_k$ is given as:

\vspace{-.45in}
\begin{eqnarray}
	Pr(\base \rightarrow \object_k) = \frac{1}{Z}\prod_{i=0}^{m}\potential(C_i),
\label{eq:joint_prob}
\end{eqnarray}
\vspace{-.2in}

\noindent where $C_i$ is defined as a clique in a fully-connected graph
that has as nodes $b_1, b_2, b_3$ and $b_4$, $Z$ is the normalization
constant and $\potential(C_i)$ corresponds to the factor potential of
the clique $C_i$.  For computational efficiency, only cliques of sizes
1 and 2 are considered, which are respectively the nodes and edges in
the complete graph of $\{b_1, b_2, b_3, b_4\}$.  The above
simplification gives rise to the following approximation of
Eqn. \ref{eq:joint_prob}:

\vspace{-.25in}
\begin{align*}
	Pr(\base \rightarrow \object_k) = \frac{1}{Z}
\prod_{i=1}^{4} \{ \potential_{node}(b_i) \prod_{j=1}^{j<i} \potential_{edge}(b_i,
b_j) \}.
\end{align*}
\vspace{-.2in}

The above operation is implemented efficiently in an incremental
manner. The last element of the implementation is the definition of
the factor potentials for nodes and edges of the graph $\{b_1, b_2,
b_3, b_4\}$. The factor potential for the nodes can be computed by
using the class probabilities returned by the CNN-based soft
segmentation, i.e.

\vspace{-.25in}
\begin{align*}
	\potential_{node}(b_i) = \pi_k(b_i).
\end{align*}
\vspace{-.2in}

The factor potentials $\potential_{edge}$ for edges can be computed using the Point-Pair Feature ({\tt PPF}) of the two points \cite{drost2010model} defining the edge and the frequency of the computed feature on the {\tt CAD} model $M_i$ of the object. The {\tt PPF} for two points on the model $m_1, m_2$ with surface normals $n_1, n_2$:

\vspace{-.45in}
\begin{align*}
	{\tt PPF}(m_1,m_2) = (\mid\mid d \mid\mid_2, \angle(n_1, d), \angle(n_2, d), \angle(n_1, n_2)),
\end{align*}
\vspace{-.3in}

\noindent wherein $d=m_2-m_1$ is the vector from $m_1$ to $m_2$.

\vspace{-.1in}
\input{select_base}
\vspace{-.1in}

A hash map is generated for the object model, which counts the number
of occurrences of discretized point pair features in the model. To
account for the sensor noise, the point pair features are
discretized. Nevertheless, even with discretization, the surface
normals of points in the scene point cloud could be noisy enough such
that they do not map to the same bin as the corresponding points on
the model. To overcome this issue, during the model generation process, each
point pair also votes to several neighboring bins. For the
accompanying implementation, the bin discretization was kept at 10
degrees and 0.5 cm. The point-pair features voted to $2^4$ other
bins in the neighborhood of the bin the feature points to. This
ensures the robustness of the method in case of noisy surface normal
computations. Then, the factor potential for edges in the base is
given as:

\vspace{-.25in}
\begin{eqnarray*}
	\potential_{edge}(b_i, b_j) = 
	\begin{cases}
	1, \text{ if } hashmap(M_k, {\tt PPF}(b_i,b_j)) > 0 \cr
	0, \text{ otherwise}
	\end{cases}
\end{eqnarray*}
\vspace{-.2in}

Thus, the sampling of bases incorporates the above definitions and
proceeds as described in Algorithm~\ref{alg:alg2}. In particular, each
of the four points in a base $B$ is sampled from the discrete
probability distribution $\pi_k$, defined for the point segment
$S_k$. This distribution is initialized as shown in
Eqns.~\ref{eq:soft_seg} and ~\ref{eq:soft_seg2} using the output of
the last layer of a \cnn. The probability of sampling a point
$p \in \segment_k$ is incrementally updated in
Algorithm~\ref{alg:alg2} by considering the edge potentials of points
with already sampled points in the base. This step essentially prunes
points that do not relate, according to the geometric model of the
object, to the already sampled points in the base. Furthermore,
constraints are defined in the form of conservative thresholds
($\epsilon_1$, $\epsilon_2$) to ensure that the selected base has a
wide interior angle and is coplanar.

The {\sc find\_congruent\_sets(B, $M_k$)} subroutine of
Algorithm~\ref{alg:alg1} is used to compute a set $\congsuperset$ of
4-points from $M_k$ that are congruent to the sampled base B. The
4-points of the base can be represented by two pairs represented by
their respective PPF and the ratio defined on the line segments by
virtue of their intersection. Two sets of point pairs are computed on
the model with the PPFs specified by the segment base. The pairs in
the two sets, which also intersect with the given ratios are
classified as congruent 4-points. The basic idea of 4 point congruent
sets was originally proposed in~\cite{aiger20084}. It was derived from
the fact that these ratios and distances are invariant across any
rigid transformation. In \stocs\ the pairs are compared using
point-pair features instead of just distances, which further reduces
the cardinality of the sets of pairs that need to be compared and thus
speed-ups the search process.
\vspace{-.1in}

%% file: compute_best_transform.tex
\begin{algorithm}[h]
\label{alg:alg1}
\caption{{\sc \stocs ($\segment_k$, $\pi_k$, $\model_k$ )} }
bestScore $\gets$ 0 \;
$\trans_{opt}$ $\gets$ identity transform\;
\While{runtime $<$ max\_runtime}{
	\base $\gets$ {\sc select\_\stocs \_base($\segment_k$, $\pi_k$, $\model_k$ )}\;
	$\congsuperset$ $\gets$ {\sc find\_congruent\_sets(\base, $\model_k$)}\;

	\ForEach{4-point set $\congset_j$ $\in \congsuperset$}
        {
		$\trans$ $\gets$ best rigid transform that aligns $\base$ to $\congset_j$ in the least squares sense\;
		score $\gets \sum_{\mpoint_i \in \model_k}f( \mpoint_i, \trans, \segment_k)$ \;
		\If{score $>$ bestScore}{
		bestScore $\gets$ score; $\trans_{opt}$ $\gets$ $\trans$ \;
		}
	}
}
return $\trans_{opt}$\;
\BlankLine
\end{algorithm}

%% file: select_base.tex
\begin{algorithm}[h]
\label{alg:alg2}
\caption{{\sc select\_\stocs \_base ($S_k$, $\pi_k$, $M_k$)} }
$b_1$ $\gets$ sample a point from $S_k$ according to the discrete probability distribution defined by the soft segmentation prior $\pi_k$ \;
\ForEach{point  $p \in S_k$}{
	$\pi(p|b_1) = \pi_k (p)   \pi_k ( b_1 )  \phi_{edge} (p, b_1)$\;
}

$b_2$ $\gets$ sample from normalized $\pi(.|b_1)$\;
\ForEach{point  $p \in S_k$}{
	$\pi(p|b_1,b_2) = \pi(p|b_1)   \pi ( b_2|b_1)  \phi_{edge} (p, b_2)$\;
	\If{$\angle((p - b_0) , (b_1-b_0)) < \epsilon_1$}{
		$\pi(p|b_1,b_2)$ $\gets$ 0 \;
	}
}
$b_3$ $\gets$ sample from normalized $\pi(.|b_1,b_2)$\;
\ForEach{point  $p \in S_k$}{
	$\pi(p|b_1,b_2,b_3) = \pi(p|b_1,b_2)   \pi ( b_3|b_1,b_2)  \phi_{edge} (p, b_3)$\;
	\If{distance(plane($b_1,b_2,b_3$), p) $<$ $\epsilon_2$}{
		$\pi(p|b_1,b_2,b_3)$ $\gets$ 0 \;
	}
}
$b_4$ $\gets$ sample from normalized $\pi(.|b_1,b_2,b_3)$\;
return $b_1, b_2, b_3, b_4$\;
\BlankLine
\end{algorithm}

%% file: 05_evaluation.tex
\vspace{-0.1in}
Two different datasets are used for the evaluation of the proposed
method.
\vspace{-0.1in}

\subsection{Amazon Picking Challenge (\apc) dataset}
\vspace{-0.1in}

This \rgbd\ dataset \cite{mitash2017improving} contains real images of
multiple objects from the Amazon Picking Challenge (\apc) in varying
configurations involving occlusions and texture-less objects.

A Fully Convolutional Network (\fcn) was trained for semantic
segmentation by using synthetic data. The synthetic images were
generated by a toolbox for this dataset \cite{mitash2017self}. A
dataset bias was observed, leading to performance drop on mean recall
for pixel-wise prediction from 95.3\% on synthetic test set to 77.9\%
on real images. Recall can be improved by using a continuous
probability output from the \fcn\ with no or very low confidence
threshold as proposed in this work. This comes at the cost of losing
precision and including parts of other objects in the segment being
considered for model registration. Nevertheless, it is crucial to achieve accurate pose estimation on real images given a
segmentation process trained only on synthetic data as it
significantly reduces labeling effort.

Table~\ref{table:apcPoseErr} provides the pose accuracy of \procs\
compared against \super\ and \vpcs. The Volumetric-4PCS (\vpcs)
approach samples 4 base points by optimizing for maximum volume and
thus coplanarity is no more a constraint. Congruency is established
when all the edges of the tetrahedron formed by connecting the points
have the same length. The performance is evaluated as mean error in
translation and rotation, where the rotation error is a mean of the
roll, pitch, and yaw error. The three processes sample 100 segment
bases and verify all the transformations extracted from the congruent
sets. While \procs\ uses soft segmentation output, the segment for the
competing approaches was obtained by thresholding on per-pixel class
prediction probability. In Table 1(a), the optimal value of the
threshold ($\epsilon =$ 0.4) is used for \super\ and \vpcs. In Figure
1(b), the robustness of all approaches is validated for different
thresholds. The percentage of successful estimates (error less than
2cm and 10 degrees) reduces with the segmentation accuracy for
both \super\ and \vpcs. But \procs\ provides robust estimates even
when the segmentation precision is very low. The \procs\ output using
FCN segmentation is comparable to results with registration on
ground-truth segmentation, which is an ideal case for the alternative
methods. This is important as it is not always trivial to compute the
optimal threshold for a test scenario.

\begin{table}[t]
\begin{center}
\begin{tabular}{p{7.5cm}p{4cm}}
\begin{tabular}{|l|c|c|c|}
\hline
Method & Rot. error & Tr. error & Time\\
\hline\hline
Super4PCS \cite{mellado2014super} & 8.83$^{\circ} $ & 1.36cm & 28.01s\\
V4PCS \cite{huang2017v4pcs} & 10.75$^{\circ}$ & 5.48cm & 4.66s\\
StoCS (OURS) & 6.29$^{\circ}$ & 1.11cm & 0.72s\\
\hline
\end{tabular}&
\raisebox{4.5\height}{\bmvaHangBox{\includegraphics[width=4cm]{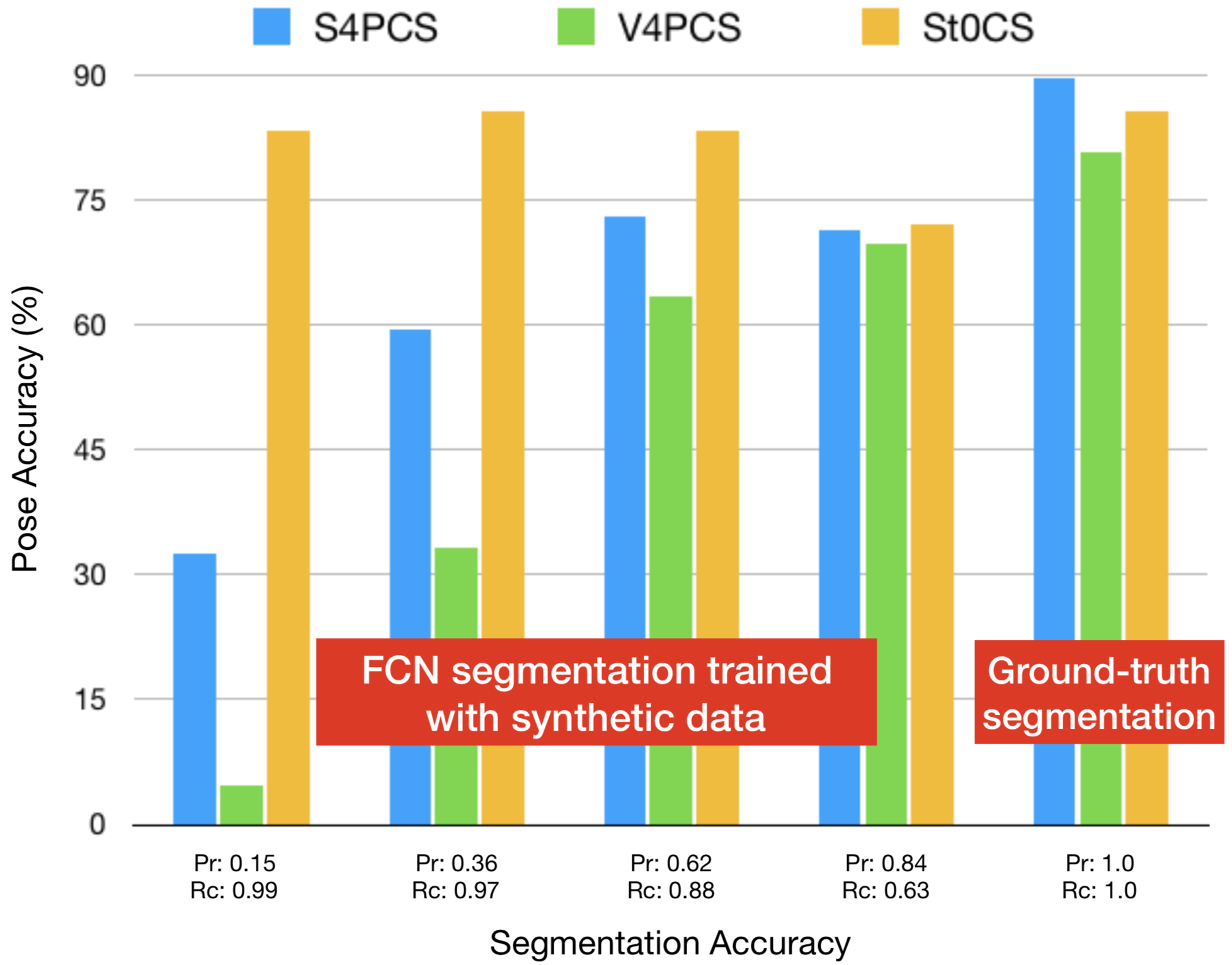}}}\\
(a) Average rotation error, translation error and execution time (per object) & (b) Robustness with varying segmentation confidences. 
\end{tabular}\\
\medskip
\begin{tabular}{c}
\begin{tabular}{|l|c|c|c|c|}
\hline
Method & Base Sampling & Set Extraction & Set Verification & \#Set per base\\
\hline\hline
Super4PCS \cite{mellado2014super} & 0.0045s & 2.43s & 19.98s & 1957.18\\
V4PCS \cite{huang2017v4pcs} & 0.0048s & 1.98s & 0.36s & 46.61\\
StoCS (OURS) & 0.0368s & 0.27s & 0.37s & 53.52\\
\hline
\end{tabular}\\
(c) Computation complexity for the different components of the registration process.
\end{tabular}
\end{center}
\vspace{-0.1in}
\caption{Comparing \stocs\ with related registration processes on the
APC dataset.}
\label{table:apcPoseErr}
\vspace{-.2in}
\end{table}

\vspace{-0.1in}
\subsection{Computational cost}
\vspace{-0.1in}

The computational cost of the process can be broken down into 3
components: base sampling, congruent set extraction, and set
verification. \procs\ increases the cost of base sampling as it
iterates over the segment to update probabilities. But this is linear
in the size of the segment and is not the dominating factor in the
overall cost. The congruent set extraction and thus the verification
step are output sensitive as the cost depends on the number of
matching pairs on the model corresponding to 2 line segments on the
sampled base for \super\ and \procs\ and 6 line segments of the
tetrahedron for \vpcs. Thus, base sampling optimizes for wide interior
angle or large volume in \super\ and \vpcs\ respectively to reduce the
number of similar sets on the model. This optimization, however, could
lead to the selection of outlier points in the sampled base, which
occurs predominantly in \vpcs. For \super\, the number of congruent
pairs still turns out to be very large (approx., 2000 per base), thus
leading to a computationally expensive set extraction and verification
stage. This is mostly seen for objects with large surfaces and
symmetric objects. \stocs\ can restrict the number of congruent sets
by only considering pairs on the model, which have the same PPF as on
the sampled base. It does not optimize for wide interior angle or
maximizing volume, but imposes a small threshold, such that nearby
points and redundant structures are avoided in base sampling. So it
can handle the computational cost without hurting accuracy as shown in
Table~\ref{table:apcPoseErr} part (c).

\vspace{-0.1in}
\subsection{\ycbvideo\ dataset}
\vspace{-0.1in}

The \ycbvideo\ dataset \cite{xiang2017posecnn} is a benchmark for
robotic manipulation tasks that provides scenes with a clutter of
21 \ycb\ objects \cite{calli2017yale} captured from multiple views and
annotated with 6-DOF object poses. Along with the dataset, the authors
also proposed an approach, \posecnn, which learns to predict the
object center and rotation solely on \rgb\ images. The poses are
further fine-tuned by initializing a modified \icp\ with the output
of \posecnn, and applying it on the depth images. The metric used for
pose evaluation in this benchmark measures the average distance
between model points transformed using the ground truth transformation
and with the predicted transform. An accuracy-threshold curve is
plotted and the area under the curve is reported as a scalar
representation of the accuracy for each approach. To ignore errors
caused due to object symmetry, the closest symmetric point is
considered as a correspondence to compute the error.

\begin{table}[t]
\begin{center}
\begin{tabular}{cc}
\begin{tabular}{|l|c|c|}
\hline
Method & Pose success & Time\\
\hline\hline
PoseCNN \cite{xiang2017posecnn} & 57.37$\%$ & 0.2s\\
PoseCNN+ICP \cite{xiang2017posecnn} & 76.53$\%$ & 10.6s\\
PPF-Hough \cite{drost2010model} & 83.97$\%$ & 7.18s\\
Super4PCS \cite{mellado2014super} & 87.21$\%$ & 43s\\
V4PCS \cite{huang2017v4pcs} & 77.34$\%$ & 4.32s\\
StoCS (OURS) & 90.1$\%$ & 0.59s\\
\hline
\end{tabular}&
\raisebox{6\height}{\bmvaHangBox{\includegraphics[width=5cm]{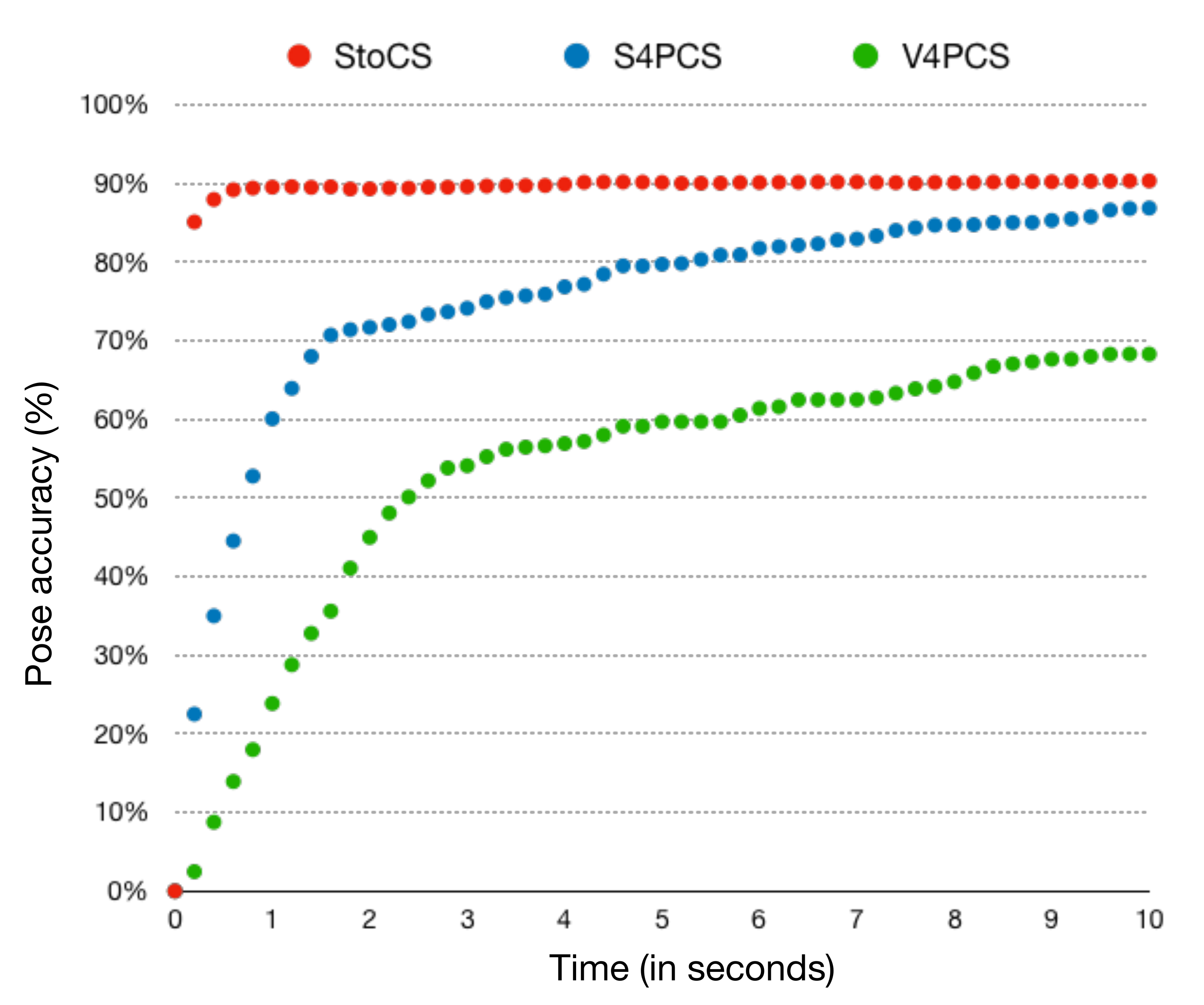}}}\\
\end{tabular}
\end{center}
\label{table:ycbeval}
\vspace{-.2in}
\caption{(left) Success given the area under the
accuracy-threshold curve and computation time (per object) on the YCB-Video
dataset. (right) Anytime results for 3 pointset registration
methods.}
\vspace{-.2in}
\end{table}

The results of the evaluation are presented in Table~2. The accuracy
of \posecnn\ is low, mostly because it does not use depth
information. When combined with a modified \icp, the accuracy
increases but at a cost of large computation time. The modified \icp\
performs a gradient-descent in the depth image by generating a
rendering score for hypothesized poses. The results are reported by
running the publicly shared code separately over each view of the
scene, which may not be optimal for the approach but is a fair
comparison point as all the compared methods are tested on the same
images.

For evaluating the other approaches, the same dataset used to train
\posecnn\ was employed to train  \fcn\ for semantic
segmentation with a {\tt VGG16} architecture. A deterministic segment
was computed based on thresholding over the network output. An
alternative that is evaluated is Hough
voting \cite{drost2010model}. This achieves better accuracy but is
computationally expensive. This is primarily due to the quadratic
complexity over the points on the segment, which perform the
voting. Next, alternative congruent set based approaches were
evaluated, \super\ and \vpcs. For each approach 100 iterations of the
algorithm were executed. As the training dataset was similar to the
test dataset, and an optimal threshold was used, 100 iterations were
enough for \super\ to find good pose estimates. Nevertheless, \super\
generates a large number of congruent sets, even when surface normals
were used to prune correspondences, leading to large computation
time. \vpcs\ achieves lower accuracy. During its base sampling
process, \vpcs\ optimizes for maximizing volume, which often biases
towards outliers.

Finally, the proposed approach was tested. A continuous soft
segmentation output was used in this case, instead of optimal
threshold and 100 iterations of the algorithm was run. It achieves the
best accuracy, and the computation time is just slightly larger
than \posecnn\ which was designed for time efficiency as it uses one
forward pass over the neural network.

\vspace{-0.1in}

%% file: 06_conclusion.tex
\vspace{-0.1in}
Scene segmentation and object pose estimation are two problems that
are frequently addressed separately.  The points provided by
segmentation are generally treated with an equal level of certainty by
pose estimation algorithms. This paper shows that a potentially better
way is to exploit the varying levels of confidence obtained from
segmentation tools, such as \cnn s. This leads to a stochastic search
for object poses that achieves improved pose estimation accuracy,
especially in setups where the segmentation is imperfect, such as when
the \cnn\ is trained using synthetic data. This is increasingly
popular for training \cnn s to minimize human
effort~\cite{Georgakis:2016aa}.


A limitation of the proposed method is the difficulty to deal with
cases where depth information is unavailable, such as with translucent
objects~\cite{Phillips:2017aa}. This can be addressed by sampling
points on hypothesized object surfaces, instead of relying fully on
points detected by depth sensors. Another extension is to generalize
the pointset bases to contain arbitrary sets of points with desirable
properties. For instance, {\it determinantal point
processes} \cite{Soshnikov:2000aa} can be used for sampling sets of
points according to their diversity.
